# Cutset Sampling with Likelihood Weighting

**Bozhena Bidyuk** and **Rina Dechter**
Information and Computer Science
University Of California Irvine
Irvine, CA 92697-3425

## Abstract

The paper extends the principle of cutset sampling over Bayesian networks, presented previously for Gibbs sampling, to likelihood weighting (LW). Cutset sampling is motivated by the Rao-Blackwell theorem which implies that sampling over a subset of variables requires fewer samples for convergence due to the reduction in sampling variance. The scheme exploits the network structure in selecting cutsets that allow efficient computation of the sampling distributions. In particular, as we show empirically, likelihood weighting over a loop-cutset (abbreviated LWLC), is time-wise cost-effective. We also provide an effective way for caching the probabilities of the generated samples which improves the performance of the overall scheme. We compare LWLC against regular liklihood-weighting and against Gibbs-based cutset sampling.

## 1 Introduction

Stochastic sampling is a popular approach for estimating answers to Bayesian queries when exact inference is intractable. Based on the generated samples, we can obtain estimates that converge to the exact values as the number of samples increases. However, convergence may be slow in large networks due to increase in sampling variance. This is the problem we address in this paper.

Based on Rao-Blackwell theorem, we can reduce sampling variance and speed up convergence by sampling only a subset of the variables (a cutset). However, the efficiency of sampling from lower-dimensional spaces is hindered by the overhead of computing the sampling distributions. The latter is equivalent to performing exact inference which is exponential in the induced width of the network whose instantiated variables (evidence and sampled) are removed.

We defined previously an efficient parametrized Gibbs cutset sampling scheme, called $w$-cutset sampling [2, 3], where the complexity of generating a single sample is bounded exponentially by $w$. In this paper, we extend the cutset sampling principle to likelihood weighting (LW) [11, 21], which is a form of importance sampling [21], focusing on sampling from a loop-cutset. The resulting scheme, which we call LWLC, computes a sample over a loop-cutset $C$ in $O((|C| + |E|) \cdot N)$, where $E$ is evidence and $N$ is the size of the input network. While we present our scheme for LW, it is applicable to other importance sampling schemes.

While both cutset schemes, one based on Gibbs sampling and one based on likelihood weighting, exploit the network structure to manage the complexity of exact inference, they compute different sampling distributions. Gibbs sampler draws a new value of variable $X_i$ from distribution $P(X_i|x \setminus x_i)$. Likelihood weighting samples a new value from $P(X_i|x_1, ..., x_{i-1})$. Furthermore, while both schemes benefit from reducing the size of the sampling space, it is hard to predict which of the two schemes is superior. The convergence speed of Gibbs sampling depends on the maximum correlation between the sampled variables. The convergence of likelihood weighting is affected by the distance between the *sampling* and the *target* distributions and, thus, depends also on the nature of evidence. Finally, Gibbs estimates converge only when all Markov Chain transition probabilities are positive.

The advantages of cutset-based importance sampling, also known as Rao-Blackwellised importance sampling, were demonstrated previously in a few special cases [9, 8, 1]. Our scheme automates the cutset selection process based on the Bayesian network structure. We demonstrate empirically that LWLC is efficient time-wise and has a lower rejection rate in networks with determinism. We achieve additional improvements by caching the probabilities of the generated samples.

Our scheme can be generalized to other importance sampling schemes.

## 2 Background

DEFINITION 2.1 (belief networks)
Let $X=\{X_1,...,X_n\}$ be a set of random variables over multi-valued domains $\mathcal{D}(X_1),...,\mathcal{D}(X_n)$. A belief network is a pair $<G,\mathcal{P}>$ where $G$ is a directed acyclic graph on $X$ and $\mathcal{P}=\{P(X_i|pa_i)|X_i \in X\}$ is the set of conditional probability tables (CPTs), conditioned on parents $pa_i$ of $X_i$. An evidence $e$ is an instantiated subset of variables $E$. A network is **singly-connected** (also called a **poly-tree**), if its underlying undirected graph has no cycles. Otherwise, it is **multiply-connected**.

DEFINITION 2.2 (loop-cutset) A **loop** in $G$ is a subgraph of $G$ whose underlying graph is a cycle. A vertex $v$ is a **sink** with respect to a loop $L$ if the two edges adjacent to $v$ in $L$ are directed into $v$. Every loop contains at least one vertex that is not a sink with respect to that loop. Each vertex that is not a sink with respect to a loop $L$ is called an **allowed** vertex with respect to $L$. A **loop-cutset** of a directed graph $G$ is a set of vertices that contains at least one allowed vertex with respect to each loop in $G$.

The queries over a singly-connected network can be processed in time linear in the size of the network [18]. In general, the complexity of queries can be reduced by restricting $G$ to a relevant subnetwork.

DEFINITION 2.3 (Relevant Subnetwork) A variable $X_i$ in DAG $G$ over $X$ is **irrelevant** (barren) w.r.t. a subset $Z \subset X$ if $X_i \notin Z$ and $X_i$ only has irrelevant descendants (if any). The relevant subnetwork of $G$ w.r.t. a subset $Z$ is the subgraph of $G$ obtained by removing all variables that are irrelevant w.r.t $Z$.

### 2.1 Likelihood Weighting

Likelihood weighting [11, 21] belongs to a family of importance sampling schemes that draw independent samples from a trial distribution $Q(X)$. The trial distribution is different from the target distribution $P(X)$. Generally, $Q(X)$ is selected so that it is easy to compute. A typical query in Bayesian networks is to estimate the posterior marginals $P(x_i|e)$ which can be obtained from the sampling estimates of $P(e)$ and $P(x_i,e)$. Let $Y = X \backslash E$. Then:

$$E_P[P(e)] = \sum_y P(y,e) = \sum_y \frac{P(y,e)}{Q(y)} Q(y) = E_Q[\frac{P(y,e)}{Q(y)}]$$

Consequently, the sampling estimate $\hat{P}(e)$ of $P(e)$, based on $T$ samples from $Q(X)$, is obtained by:

$$\hat{P}(e) = \frac{1}{T} \sum_{t=1}^T \frac{P(y^{(t)},e)}{Q(y^{(t)})} = \frac{1}{T} \sum_{t=1}^T w^{(t)} \quad (1)$$

where $w^{(t)} = \frac{P(y^{(t)},e)}{Q(y^{(t)})}$ is the *weight* of sample $y^{(t)}$. In a similar manner, but counting only those samples where $X_i = x_i$, we can obtain an expression for the sampling estimate $\hat{P}(x_i,e)$ of $P(x_i,e)$ for $X_i \in X\backslash E$ by:

$$\hat{P}(x_i,e) = \frac{1}{T} \sum_{t=1}^T w^{(t)} \delta(x_i, x^{(t)}) \quad (2)$$

where $\delta(x_i, x^{(t)})=1$ iff $x_i^{(t)}=x_i$ and $\delta(x_i, x^{(t)})=0$ otherwise. Since $P(x_i|e) = \frac{P(x_i,e)}{P(e)}$, we get:

$$\hat{P}(x_i|e) = \frac{\sum_{t=1}^T w^{(t)} \delta(x_i, x^{(t)})}{\sum_{t=1}^T w^{(t)}} = \alpha \sum_{t=1}^T w^{(t)} \delta(x_i, x^{(t)}) \quad (3)$$

where $\alpha$ is a normalization constant. These sampling estimates are guaranteed to converge to their target values as $T$ increases as long as the *support* for $Q(X)$ includes all support for $P(X)$. Namely, the condition $\forall x \in X, P(x) \neq 0 \Rightarrow Q(x) \neq 0$ must hold. Eq. (3) yields a *biased* estimate of $\hat{P}(x_i|e)$. However, when the sample size is large enough, bias can be ignored [10].

Likelihood weighting draws samples from a distribution $Q(X)$ that is close to the prior distribution. It begins with a network without evidence and assigns values to nodes in topological order. First, root nodes are sampled from their prior distributions. Then, the values of all other nodes $X_i \in X\backslash E$ are sampled from the distribution $P(X_i|pa_i)$. Evidence variables $E_i \in E$ are assigned their observed value. Thus, the sampling distribution of likelihood weighting can be described as follows:

$$Q(X) = \prod_{X_i \in X\backslash E} P(X_i|pa_i) |_{E=e} \quad (4)$$

We therefore compute the weight $w^{(t)}$ of sample $t$ by:

$$w^{(t)} = \frac{P(x^{(t)})}{Q(x^{(t)})} = \frac{\prod_{X_i \in X} P(x_i^{(t)}|pa_i^{(t)})}{\prod_{X_i \in X\backslash E} P(x_i^{(t)}|pa_i^{(t)})} \quad (5)$$

All factors in the numerator and denominator of the fraction cancel out except for $P(e_i|pa_i)$, leaving:

$$w^{(t)} = \prod_{E_i \in E} P(e_i|pa_i^{(t)}) \quad (6)$$

Thus, during sampling, we compute the weight $w^{(t)}$ of sample $t$ by initializing $w^{(t)} \leftarrow 1$ and updating $w^{(t)} \leftarrow w^{(t)} \cdot P(e_i|pa_i^{(t)})$ whenever we encounter an evidence $E_i = e_i$. The posterior marginals estimates are obtained by plugging the sample weights in Eq.(3).

The convergence of importance sampling schemes can be slow when $Q(X)$ is very different from $P(X)$. Consequently, many importance sampling schemes focus on finding an improved sampling distribution by either changing the variable sampling order [12] or updating the sampling distribution based on previously generated samples [21, 5, 22]. We can also improve convergence by reducing the dimensionality of the sampling space as implied by Rao-Blackwell theorem.

## 3 Rao-Blackwellised Likelihood Weighting

Give a Bayesian network over a set of variables $X$ with evidence $E \subset X$, $E=e$, let $C \subset X \backslash E$ be a subset of variables in $X$, $Z = C \bigcup E$, and $m=|Z|$. Let $o=\{Z_1,...,Z_m\}$ be a topological ordering of the variables. We can define likelihood weighting over $Z$ as follows. Processing variables in order $o$, we sample value $z_1$ from distribution $P(Z_1)$, $z_2$ from $P(Z_2|z_1)$, and so on. For each $Z_i \in C$, we sample a value $z_i$ from the distribution $P(Z_i|z_1,...,z_{i-1})$. If $Z_i \in E$, we assign $Z_i$ its observed value. The sampling distribution $Q(C)$ is:

$$Q(C) = \prod_{Z_i \in C} P(Z_i|z_1,...,z_{i-1})|_{E=e} \quad (7)$$

The weight $w^{(t)}$ of sample $t$ is given by:

$$w^{(t)} = \frac{P(z^{(t)})}{Q(z^{(t)})} = \frac{\prod_{Z_i \in Z} P(z_i^{(t)}|z_1^{(t)},...,z_{i-1}^{(t)})}{\prod_{Z_i \in Z \backslash E} P(z_i^{(t)}|z_1^{(t)},...,z_{i-1}^{(t)})} \quad (8)$$

After cancelling out the common factors in denominator and numerator, we get:

$$w^{(t)} = \prod_{Z_i \in E} P(e_i|z_1^{(t)},...,z_{i-1}^{(t)}) \quad (9)$$

During sampling, the weight (initialized to 1) is updated every time we encounter an evidence variable $Z_i \in E$ with observed value $e_i$ using:

$$w^{(t)} \leftarrow w^{(t)} \cdot P(e_i|z_1,...,z_{i-1}) \quad (10)$$

The main difference between likelihood weighting over cutset $C$ and sampling over all variables $X$ is in computing the sampling distributions. In the latter case, the distribution $P(X_i|x_1,...,x_{i-1}) = P(X_i|pa_i)$ is readily available in the conditional probability table of $X_i$. However, the sampling distribution $P(Z_i|z_1,...,z_{i-1})$ for LWLC needs to be computed.

Consider the special case when $C \cup E$ is a loop-cutset. In this case, we can compute the probability $P(z)=P(c,e)$ in linear time and space using Pearl's belief propagation algorithm. We can show that we can also compute $P(Z_i|z_1,...,z_{i-1})$ efficiently if we order the variables in $Z$ topologically and restrict our attention to the relevant subnetwork of $Z_1,...,Z_i$.

THEOREM 3.1 *Given Bayesian network over $X$, evidence $E \subset X$, and cutset $C \subset X \backslash E$, let $Z = C \cup E$ be a loop-cutset. If $Z$ is topologically ordered, then $\forall Z_j \in Z$ the relevant subnetwork of $Z_1,...,Z_j$ is singly-connected when $Z_1,...,Z_j$ are observed.*

**Proof.** Proof by contradiction. Assume that the relevant subnetwork of $Z_1,...,Z_j$ contains a loop $L$ with sink $S$. Then, either $S=Z_q$ or $S$ has a descendant $Z_q$, $1 \leq q \leq j$, (otherwise $S$ is irrelevant). By definition of loop-cutset, $\exists C_m \in L$ s.t. $C_m \neq S$ and $C_m \in C \subset Z$. Therefore, $C_m$ is an ancestor of $Z_q$. Since variables are topologically ordered and all loop-cutset nodes preceding $Z_q$ are observed, $C_m$ must be observed, thus, breaking the loop, yielding a contradiction. □

Conclusion: if $C$ is a loop-cutset, we can compute the distributions $P(Z_i|z_1,...,z_{i-1})$ for every $Z_i \in Z$ over the relevant subnetwork of $Z_i$ in linear time and space.

Therefore, the complexity of computing a new sample is proportional to the number of variables in $Z$ and the size of the input $N$. In summary:

THEOREM 3.2 (**Complexity**) *Given a Bayesian network over $X$, evidence $E$, and a loop-cutset $C \subset X \backslash E$, the complexity of generating one sample using likelihood weighting over a cutset $C$ is $O(|Z| \cdot N)$ where $Z = C \cup E$ and $N$ is the size of the input network.*

Once a sample $c^{(t)}$ is generated, we apply belief propagation algorithm one more time to obtain the posterior marginals, $P(X_i|c^{(t)}, e)$, for each remaining variable. Once $T$ samples are generated, we obtain the posterior marginals estimates, similar to Eq. (3), by:

$$\hat{P}(c_i|e) = \alpha \sum_{t=1}^{T} w^{(t)} \delta(c_i, c^{(t)}), \forall C_i \in C$$

$$\hat{P}(x_i|e) = \alpha \sum_{t=1}^{T} w^{(t)} P(x_i|c^{(t)}, e), \forall X_i \in X \backslash C, E$$

### 3.1 Convergence

Likelihood weighting on a loop-cutset (LWLC) has a higher overhead in computing the distributions $P(Z_i|z_1,...,z_{i-1})$ for $\forall Z_i \in Z$, compared with sampling on a full variable set. However, as mentioned earlier, it converges faster. In general, importance sampling convergence rate is affected by the sampling variance and the distance between the sampling and the target distributions. The estimates obtained by sampling from a lower-dimensional space have lower variance due to Rao-Blackwell theorem. That is:

$$Var\{\frac{P(Y,C)}{Q(Y,C)}\} \geq Var\{\frac{P(C)}{Q(C)}\}$$

where $P(C) = \sum_y P(Y,C)$ and $Q(C) = \sum_y Q(Y,C)$ [9, 16] A proof can be found in [9] and [16]. Consequently, fewer LWLC samples are needed to achieve the same accuracy as LW.

The information distance between target distribution $P(C|e)$ and sampling distribution $Q(C)$ in LWLC is smaller than the distance between $P(X|e)$ and sampling distribution $Q(X)$. We can show this for the KL-distance [15]:

$$KL(P(X), Q(X)) = \sum_x P(x) \log \frac{P(x)}{Q(x)} \quad (11)$$

THEOREM **3.3 (Reduced Information Distance)**
*Given a Bayesian network expressing probability distribution $P(X)$, evidence $E=e$, and a cutset $C \subset X \backslash E$, let $Q(X)$ and $Q(C,E)$ denote the likelihood weighting sampling distribution over $X$ and over $C,E$ respectively. Then:*

$$KL(P(C|e), Q(C,E)) \leq KL(P(X|e), Q(X))$$

We outline the proof in the Appendix. The details are available in [4].

### 3.2 Caching Sampling on a Cutset

Often, we can reduce the computation time of a sampling scheme by caching the generated samples and their probabilities. Caching LW values is of limited benefit since it uses probabilities stored in CPTs. However, in the case of LWLC, caching may compensate in part for the computation overhead. A suitable data structure for caching is a search-tree over the cutset $C$ with a root node $C_1$. As new variable values are sampled and a partial assignment to the variables $C_1, ..., C_i$ is generated, LWLC traverses the search tree along the path $c_1, ..., c_i$. Whenever a new value of $C_i$ is sampled, the corresponding tree branch is expanded and the current sample weight and the sampling distribution $P(C_i|z_1, ..., z_{i-1})$ are saved in the node $C_i$. In the future, when generating the same partial assignment $c_1, ..., c_i$, LWLC saves on computation by reading saved distributions from the tree. We will use LWLC-BUF to denote LWLC sampling scheme that uses a memory buffer to cache previously computed probabilities. LWLC-BUF can also update the sampling distributions $P(C_i|z_1, ..., z_{i-1})$ when dead-ends are discovered. Namely, if the algorithm finds that a partial instantiation $z_1, ..., z_i$, cannot be extended to a full tuple with non-zero probability, then we set $P(C_i|z_1, ..., z_{i-1}) = 0$ and normalize the updated distribution.

## 4 Experiments

### 4.1 Methodology

In this section, we compare empirically the performance of full likelihood weighting (LW), sampling over all the variables, against likelihood weighting on a loop-cutset (LWLC) and buffered likelihood weighting on a loop-cutset (LWLC-BUF). In networks with positive distributions, we compare likelihood weighting side by side with Gibbs sampling (Gibbs) and Gibbs-based loop-cutset sampling (LCS) [2]. For reference, we also compare with the estimates obtained by Iterative Belief Propagation (IBP). Belief propagation computes the exact posterior marginals in poly-trees [18]. When applied to networks with loops, it computes approximate marginals when it converges. IBP is fast and often produces good estimates [17, 20].

The quality of the approximate posterior marginals is measured by the Mean Square Error (MSE):

$$MSE = \frac{\sum_{X_i \in X \backslash E} \sum_{\mathcal{D}(X_i)} [P(x_i|e) - \hat{P}(x_i|e)]^2}{\sum_{X_i \in X \backslash E} |\mathcal{D}(X_i)|}$$

The exact posterior marginals $P(X_i|e)$ are obtained by bucket-tree elimination [7, 6]. We also measure the rejection rate $R$ of each sampling scheme.

Table 1: Benchmarks' characteristics: $N$-number of nodes, $w^*$-induced width, $|LC|$-loop-cutset size, $P(e)$-average probability of evidence (over 30 instances), $T_{BE}$-exact computation time by bucket elimination.

|  | **N** | **w\*** | **\|LC\|** | **P(e)** | **T$_{BE}$** |
|---|---|---|---|---|---|
| cpcs360b | 360 | 21 | 26 | 5E-8 | 20 min |
| cpcs422b | 422 | 22 | 47 | 1.5E-6 | 50 min |
| Pathfinder1 | 109 | 6 | 9 | 0.07 | 1 sec |
| Pathfinder2 | 135 | 4 | 4 | 0.06 | 0.01 sec |
| Link | 724 | 15 | 142 | 0.07 | 325 sec |

Our benchmarks are taken from Bayesian network repository. They include two subsets of Pathfinder network, Pathfinder1 and Pathfinder2, Link, and two CPCS networks, cpcs360b and cpcs422b. The benchmarks' properties are summarized in Table 1. Pathfinder is an expert system for identifying disorders from lymph node tissue sections [13]. Link is a model for the linkage between two genes [14]. The exact posterior marginals for those networks were easy to compute by bucket elimination. However, they are hard for sampling because of the large number of deterministic relationships. cpcs360b and cpcs422b are derived from the Computer-Based Patient Care Simulation system [19]. They are more challenging for exact inference because of their large induced widths. All experiments were performed on a 1.8 GHz CPU.

## 4.2 Results

### 4.2.1 Sampling Speed

We generated 30 instances of each network with different random observations among the leaf nodes. In Table 2, we report the speed of generating samples using LW, LWLC, and LWLC-BUF sampling schemes. As expected, LWLC generates far fewer samples than LW. Notably, the relative speed of LW and LWLC remains the same in the two Pathfinder networks and in Link network. By the time LW generates $100,000$ samples, LWLC generates 1200 samples. Table 2 also shows an order of magnitude improvement in the speed of generating samples by LWLC-BUF in cpcs360b, Pathfinder1, and Pathfinder2, a factor of 2 improvement in cpcs422b, and no change in the Link network. The improvement depends on the ratio of unique samples. The number of unique tuples in Pathfinder networks is only $\approx 1\%$ of the total number of samples and, thus, 99% of the computation is redundant. However, in Link network, nearly all samples are unique. Hence, buffering was not beneficial.

Table 2: Average # of samples generated by LWLC and LWLC-BUF by the time LW generates $100,000$ samples.

|            | LW     | LWLC | LWLC-BUF |
|------------|--------|------|----------|
| cpcs360b   | 100000 | 2400 | 24000    |
| cpcs422b   | 100000 | 25   | 50       |
| Pathfinder1| 100000 | 1200 | 12000    |
| Pathfinder2| 100000 | 1200 | 12000    |
| Link       | 100000 | 1200 | 1200     |

### 4.2.2 Rejection Rates

Table 3: Average rejection rates for different benchmarks: **k** -# instances, out of 30, where rejection rate <100%, **R** - average rejection rate.

|      | LW  |       | LWLC |       | LW-BUF |       |
|------|-----|-------|------|-------|--------|-------|
|      | k   | R(%)  | k    | R(%)  | k      | R(%)  |
| PF1  | 30  | 47    | 30   | 6     | 30     | 0.01  |
| PF2  | 28  | 77    | 30   | 26    | 30     | 0.05  |
| Link | 17  | 67    | 30   | 16    | 30     | 16    |

When target distribution $P(X)$ has many zeros where sampling distribution $Q(X)$ remains positive, many samples with weight 0 may be generated which do not contribute to the sampling estimates. Hence, we call them "rejected." This is not an issue in cpcs360b and cpcs422b where all probabilities are positive. However, in deterministic networks, many samples may be rejected, contributing to slow convergence. We will use the *rejection rate* $R$ to denote the percentage of samples of weight 0. When the evidence is rare, we may need to generate a very large number of samples before we find a single sample of non-zero weight. When all samples are rejected, we will say that the rejection rate is 100% and call the network instance *unresolved*.

The rejection rates of the three likelihood weighting schemes over Pathfinder1, Pathfinder2, and Link are summarized in Table 3. For each benchmark, we report the number of instances $k$ (out of 30), where the rejection rate <100%. As we can see, LW resolved all 30 instances of Pathfinder1 but only 28 instances of Pathfinder2 and only 17 instances of Link. LWLC and LWLC-BUF resolved all network instances.

Table 3 also reports the rejection rate $R$ averaged over those instances where all three algorithms generated some samples with non-zero probabilities. As we can see, LW has high rejection rates in all benchmarks. The corresponding LWLC rejection rates are a factor of 3 or more smaller. Although lower rejection rate alone does not guarantee faster convergence, it helps compensate for generating fewer samples. The rejection rate of LWLC-BUF is two orders of magnitude lower than LWLC in Pathfinder networks but it is the same as LWLC in Link network (also because most of the samples are unique).

The rejection rate of LW and LWLC does not change with time. However, as LWLC-BUF learns zeros of the target distribution, its rejection rate may decrease as the number of samples increases. Figure 1 demonstrates this on the example of Pathfinder networks.

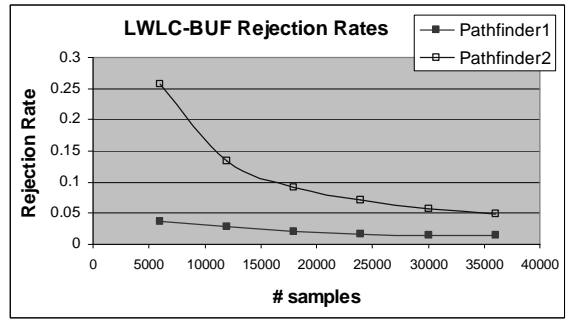

Figure 1: LWLC-BUF average rejection rate over 30 network instances in Pathfinder1 and Pahfinder2 as a function of the number of samples.

### 4.2.3 Accuracy of the Estimates

The MSE results for PathFinder1, Pathfinder2, and Link are shown in Figure 2 as a function of time. The comparative behavior of LW, LWLC, and LWLC-BUF sampling schemes is similar in all three networks. LWLC consistently converges faster than LW and out-performs IBP within 2 seconds. LW outperforms IBP within 2 seconds in Pathfinder1 and within 8 seconds in Pathfinder2. However, LW is considerably worse than IBP in Link network. LWLC-BUF converges faster than LWLC in Pathfinder1 and Pathfinder2 because it generates more samples and has a lower rejection rate. In Link network, their performance is the same and, thus, we only show the LWLC curve.

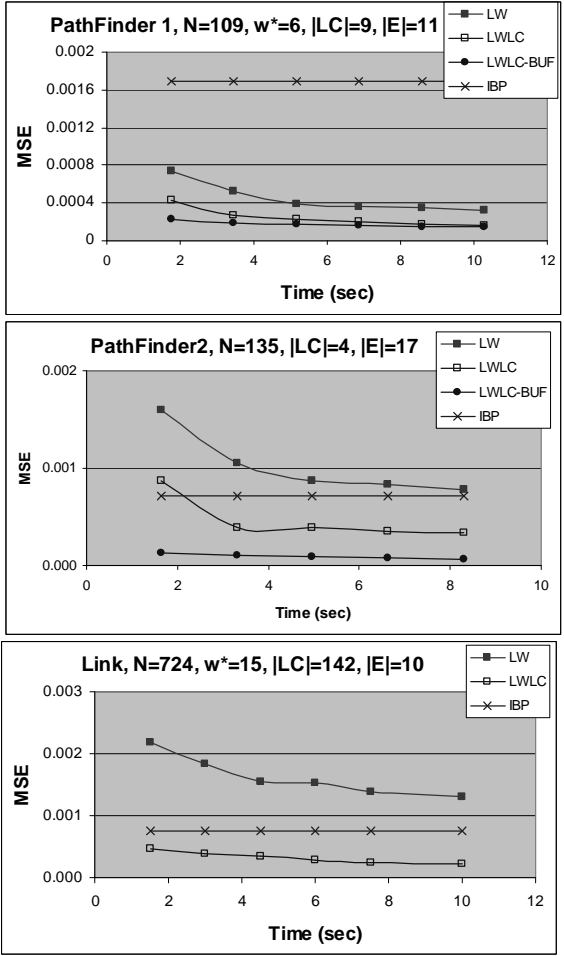

Figure 2: MSE as a function of time for LW, LWLC, LWLC-BUF, and IBP over 30 network instances of Pathfinder1 (top), 28 instances of Pathfinder2 (middle), and 17 instances of Link (bottom).

The PathFinder2 network was also used as a benchmark in the evaluation of AIS-BN algorithm [5], an adaptive importance sampling scheme. Although we experimented with different network instances, we can make a rough comparison. Within 60 seconds, AIS-BN computes MSE $\approx$ 0.0005. Adjusting for the difference in processor speed, the corresponding MSEs of LWLC and LWLC-BUF are $\approx$0.004 and $\approx$0.00008, obtained in 6 seconds. Hence, AIS-BN and LWLC-BUF produce comparable results.

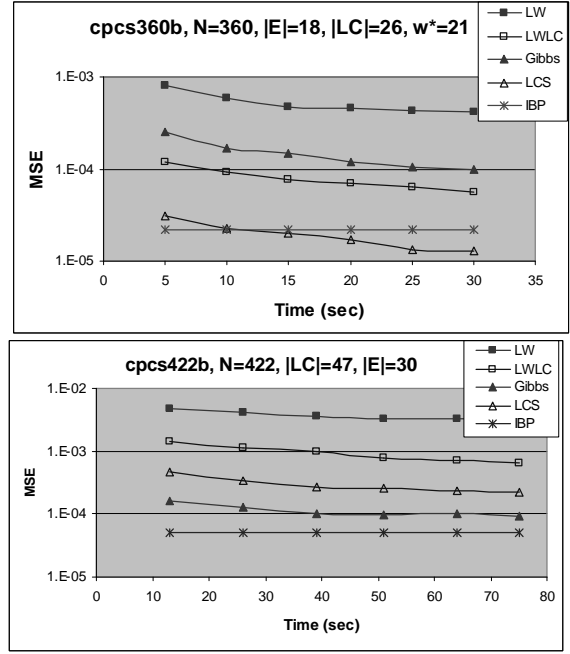

Figure 3: MSE as a function of time for full Gibbs sampling (Gibbs), Gibbs loop-cutset sampling (LCS), LW, LWLC, and IBP in cpcs360b and cpcs422b.

The accuracy of LW and LWLC for cpcs360b and cpcs422b networks is shown in Figure 3. Overall, results are similar. The LWLC outperforms LW by a wide margin in both benchmarks. Since all probabilities are positive, we also show the results for two Gibbs sampling schemes. Gibbs outperforms full likelihood weighting. Gibbs-based loop-cutset sampling (LCS) outperforms LWLC. Figure 4 focuses on the buffered cutset sampling schemes. Both LWLC-BUF and LCS-BUF improve substantially over the plain LWLC and LCS. And again, the Gibbs-based LCS-BUF is better than LWLC-BUF.

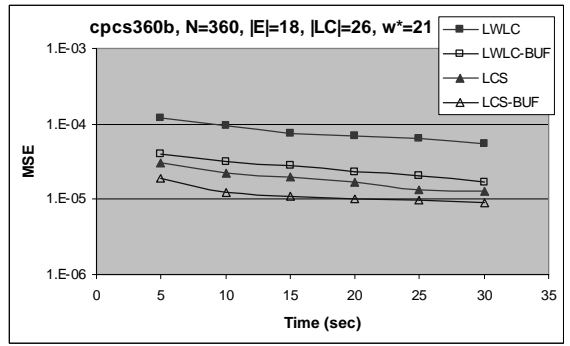

Figure 4: MSE in cpcs360b as a function of time for LCS and LWLC vs. buffered LCS and LWLC-BUF.

Although Gibbs sampling schemes outperformed likelihood weighting methods in cpcs360b and cpsc422b, where evidence was selected among leaf nodes, the two methods are likely to switch places when fewer leaf nodes are observed. In particular, likelihood weighting outperforms Gibbs sampling in cpcs360b and cpcs422b without evidence [4].

## 5 Related Work and Conclusions

In this paper we presented a cutset-based likelihood weighting. By reducing the dimensionality of the sampling space, we achieve reduction in th esampling variance and also reduce the information distance (KL-distance) between the sampling and the target distributions. Therefore, the cutset sampling scheme requires fewer samples to converge.

In the past, Rao-Blackwellised importance sampling was made efficient by exploiting the properties of the conditional probability distributions, e.g., when the distributions for the marginalised variables could be computed analytically using a Kalman filter [9, 8, 1] or when the marginalised variables in a factored HMM became conditionally independent (when sampled variables are observed) due to the numerical structure of the CPTs [8]. In contrast, our method bounds the complexity of computing the sampling distributions by exploiting the structure of the network.

We demonstrated empirically that cutset-based likelihood weighting is time-wise effective. Namely, it computes more accurate estimates than likelihood weighting as a function of time. We improve the convergence of cutset-based likelihood weighting by caching previously computed samples. The buffered scheme reduces the average sample computation time since it does not re-compute the probabilities of previously generated tuples and since it allows modifying the cached distributions dynamically.

In this paper, we only updated the saved distributions when a partially-instantiated cutset tuple could not be extended to a full cutset tuple with non-zero probability. However, we can additionally update cached distributions based on the weight of previously generated samples as adaptive importance sampling techniques do. The proposed cutset-based likelihood weighting can be generalized to other importance sampling schemes.

## Appendix

Here, we outline the proof for Theorem 3.3. To simplify notation, we let $KL_x = KL(P(Y,C|e), Q(Y,C))$ and $KL_c = KL(P(C|e), Q(C))$. By definition:

$$KL_c = \sum_c P(c|e) \lg \frac{P(c|e)}{Q(c)} \quad (12)$$

$$KL_x = \sum_{y,c} P(y,c|e) \lg \frac{P(y,c|e)}{Q(y,c)} \quad (13)$$

The proof consists of several transformation steps. First, for $KL_c$, we replace the conditional probability $P(c|e)$ in the numerator of the fraction with $\frac{P(c,e)}{P(e)}$. Then, we replace $\lg \frac{P(c,e)}{Q(c)Q(e)}$ with $\lg \frac{P(c,e)}{Q(c)} - \lg P(e)$ and open parenthesis, yielding:

$$KL_c = \sum_c P(c|e) \lg \frac{P(c,e)}{Q(c)} - \sum_c P(c|e) \lg P(e) \quad (14)$$

Factoring out $\lg P(e)$ from the sum, we sum out $\sum_c P(c|e) = 1$ and get:

$$KL_c = \sum_c P(c|e) \lg \frac{P(c,e)}{Q(c)} - \lg P(e) \quad (15)$$

We apply similar transformations to $KL_x$ and obtain:

$$KL_x = \sum_{y,c} P(y,c|e) \lg \frac{P(y,c,e)}{Q(y,c)} - \lg P(e) \quad (16)$$

To simplify the analysis, we denote:

$$KL'_c = \sum_c P(c|e) \lg \frac{P(c,e)}{Q(c)} \quad (17)$$

$$KL'_x = \sum_{y,c} P(y,c|e) \lg \frac{P(y,c,e)}{Q(y,c)} \quad (18)$$

Since $KL_x - KL_c = KL'_x - KL'_c$, we continue analysis of $KL'_x$ and $KL'_c$ only. Using $c_{1:k_i}$ and $e_{1:i-1}$ to denote respectively the subsets of cutset variables and evidence variables that precede variable $E_i \in E$ in sampling order, we replace $\frac{P(y,c,e)}{Q(y,c)} = \prod_{E_i} P(e_i|pa_i)$ and $\frac{P(c,e)}{Q(c)} = \prod_{E_i} P(e_i|c_{1:k_i}, e)$, yielding:

$$KL'_c = \sum_c P(c|e) \lg \prod_{E_i} P(e_i|c_{1:k_i}, e) \quad (19)$$

$$KL'_x = \sum_{y,c} P(y,c|e) \lg \prod_{E_i} P(e_i|pa_i) \quad (20)$$

Since log of a product equals the sum of logs, we obtain:

$$KL'_c = \sum_c P(c|e) \sum_{E_i} \lg P(e_i|c_{1:k_i}, e) \quad (21)$$

$$KL'_x = \sum_{y,c} P(y,c|e) \sum_{E_i} \lg P(e_i|pa_i) \quad (22)$$

Next, we sum out variables that do not appear under the log function:

$$KL'_c = \sum_{E_i} \sum_{c_{1:k_i}} P(c_{1:k_i}|e) \lg P(e_i|c_{1:k_i}, e)$$

$$KL'_x = \sum_{E_i} \sum_{c_{1:k_i}} \sum_{pa_i} P(c_{1:k_i}|e) P(pa_i|c_{1:k_i}, e) \lg P(e_i|pa_i)$$

Using Jensen's inequality, we get a lower bound on $KL'_x$:

$$KL'_x \geq \sum_{E_i} \sum_{c_{1:k_i}} P(c_{1:k_i}|e) \lg \sum_{pa_i} P(e_i|pa_i) P(pa_i|c_{1:k_i}, e)$$

Then, we evaluate the difference $\Delta = KL'_x - KL'_c$ between the two KL-distances:

$$\Delta \geq \sum_{E_i} \sum_{c_{1:k_i}} P(c_{1:k_i}|e) \lg \sum_{pa_i} P(e_i|pa_i) P(pa_i|c_{1:k_i}, e)$$
$$- \sum_{E_i} \sum_{c_{1:k_i}} P(c_{1:k_i}|e) \lg P(e_i|c_{1:k_i}, e)$$
$$= \sum_{E_i} \sum_{c_{1:k_i}} P(c_{1:k_i}|e) \lg \frac{\sum_{pa_i} P(e_i|pa_i) P(pa_i|c_{1:k_i}, e)}{P(e_i|c_{1:k_i}, e)}$$

Finally, we show that the value of the expression under the log is $\geq 1$ and, thus, the log value is positive. Consequently, $KL_x - KL_c = KL'_x - KL'_c \geq 0$.